\documentclass[runningheads]{llncs}
\usepackage[T1]{fontenc}

\usepackage{graphicx,verbatim}
\usepackage{booktabs}
\usepackage{multirow}
\usepackage{pifont}
\usepackage{bm}
\usepackage[table]{xcolor}
\usepackage{amsmath}
\usepackage{soul}
\usepackage{caption} 
\usepackage{amssymb}
\let\Letter\relax
\usepackage[misc]{ifsym}
\usepackage{xurl}
\usepackage{hyperref}
\hypersetup{
    colorlinks=true,    
    urlcolor=blue,      
    linkcolor=blue,     
    citecolor=blue      
}
\newcommand{\blue}[1]{\textcolor{blue}{#1}}
\makeatletter
\renewcommand\@makefnmark{}
\makeatother

\begin{document}

\title{VIHD: Visual Intervention-based Hallucination Detection for Medical Visual Question Answering}
\titlerunning{VIHD}

\author{Jiayi Chen\inst{1}\textsuperscript{(\Letter)}\thanks{Corresponding author: J. Chen.} \and
Benteng Ma\inst{4} \and
Zehui Liao\inst{1} \and 
Winston Chong\inst{2,3} \and 
Yasmeen George\inst{1} \and 
Jianfei Cai\inst{1}
}


\authorrunning{J. Chen et al.}

\institute{Department of Data Science \& AI, Faculty of Information Technology, Monash University, Melbourne, VIC 3800, Australia
\and
Alfred Health Radiology, Alfred Health, Melbourne, VIC 3004, Australia 
\and
School of Translational Medicine, Faculty of Medicine, Nursing and Health
Sciences, Monash University, Melbourne, VIC 3800, Australia
\and
Hong Kong Polytechnic University, Hong Kong SAR, China
\\
\email{jiayi.chen@monash.edu}
}

\maketitle              

\begin{abstract}
While medical Multimodal Large Language Models (MLLMs) have shown promise in assisting diagnosis, they still frequently generate hallucinated responses that appear linguistically plausible but lack visual evidence. Such hallucinations pose risks to clinical decision-making and necessitate effective detection. Existing introspective detection methods primarily perform uncertainty estimation or logical verification by analyzing model responses conditioned on original or perturbed inputs. However, such external perturbations are often heuristic and context-agnostic, which overlooks the internal cross-modal dependency between generated tokens and related visual tokens during decoding. To address this issue, we propose VIHD, a \textbf{V}isual \textbf{I}ntervention-based \textbf{H}allucination \textbf{D}etection method that leverages targeted visual token masking to calibrate semantic entropy for more effective hallucination detection. VIHD locates visually dominant decoder layers via \textit{Visual Dependency Probing (VDP)}, executes \textit{Visual Intervention Decoding (VID)} via token masking to calibrate the semantic distribution, and quantifies the resulting \textit{Calibrated Semantic Entropy (CSE)} as a reliable hallucination signal. Extensive experiments on three medical VQA benchmarks with two medical MLLMs demonstrate that VIHD consistently outperforms state-of-the-art methods, underscoring the importance of fine-grained visual dependency for hallucination detection. The code will be available at \url{https://github.com/Jiayi-Chen-AU/VIHD}.


\end{abstract}

\section{Introduction}
Medical multimodal large language models (MLLMs) have shown remarkable efficacy in assisting medical diagnosis, supporting visual question answering (VQA) and automated report generation~\cite{li2023llava,sellergren2025medgemma,jiang2025hulu,xu2025lingshu}. However, these models are susceptible to \textit{hallucinated responses} that are linguistically coherent yet contradict visual evidence~\cite{liao2025vision,jin2026v}. This issue poses risks to clinical decision-making and underscores the need for effective hallucination detection.

Existing hallucination detection methods can be broadly categorized into three paradigms: supervised detection, external verification, and introspective detection. \textbf{Supervised methods}~\cite{gunjal2024detecting,chen2024detecting,xiao2025detecting,prabhakaran2025vade} typically train auxiliary detectors on annotated hallucination datasets. While effective in-distribution, they hinge on labor-intensive annotations and generalize poorly to novel domains or architectures. \textbf{External verification}~\cite{cohen2023lm,zhang2023sac3,yu2024mm,sunredeep} leverages auxiliary LLMs, MLLMs, or knowledge bases to validate model outputs, but introduces substantial computational overhead and relies heavily on external tools. In contrast, \textbf{introspective approaches}~\cite{li2024reference,farquhar2024detecting,liao2025vision,wu2024logical,jin2026v} are entirely self-contained, requiring neither extra training nor external resources, making them efficient and deployment-friendly. Existing introspective approaches typically estimate token uncertainty~\cite{li2024reference,ma2025estimating}, semantic uncertainty~\cite{farquhar2024detecting,zhang2024vl,liao2025vision}, or logical consistency~\cite{wu2024logical,jin2026v} by analyzing model responses to original or perturbed inputs. However, such external perturbations are inherently heuristic and context-agnostic~\cite{huoself}. They treat MLLMs as black boxes, failing to probe the internal causal dependencies between generated tokens and visual evidence during decoding.

To mitigate this issue, we propose VIHD, a training-free \textbf{V}isual \textbf{I}ntervention-based \textbf{H}allucination \textbf{D}etection method, that probes the fidelity of cross-modal reasoning through targeted token-level interventions. VIHD operates in three stages. First, we identify visually dominant layers in the LLM decoder where cross-modal attention is most prominent. Secondly, we introduce a visual intervention decoding scheme that masks high-attention visual tokens to produce intervened responses, thereby revealing the model's dependency on specific visual evidence. Third, we design an adaptive strategy that integrates original and intervened semantic distributions via complementary or contrastive fusion, yielding a calibrated semantic entropy metric for hallucination detection.

Our main contributions are threefold:
(1) We introduce VIHD, a training-free framework that harnesses fine-grained visual interventions to probe internal causal dependencies for hallucination detection.
(2) We design an adaptive pipeline that identifies critical decoder layers, performs attention-guided visual token masking, and derives calibrated semantic entropy to quantify hallucination severity.
(3) Extensive experiments on three medical VQA benchmarks and two medical MLLMs demonstrate the superiority of VIHD in hallucination detection.

\section{Methodology}
\begin{figure}[tbp]
    \centering
    \includegraphics[width=\linewidth]{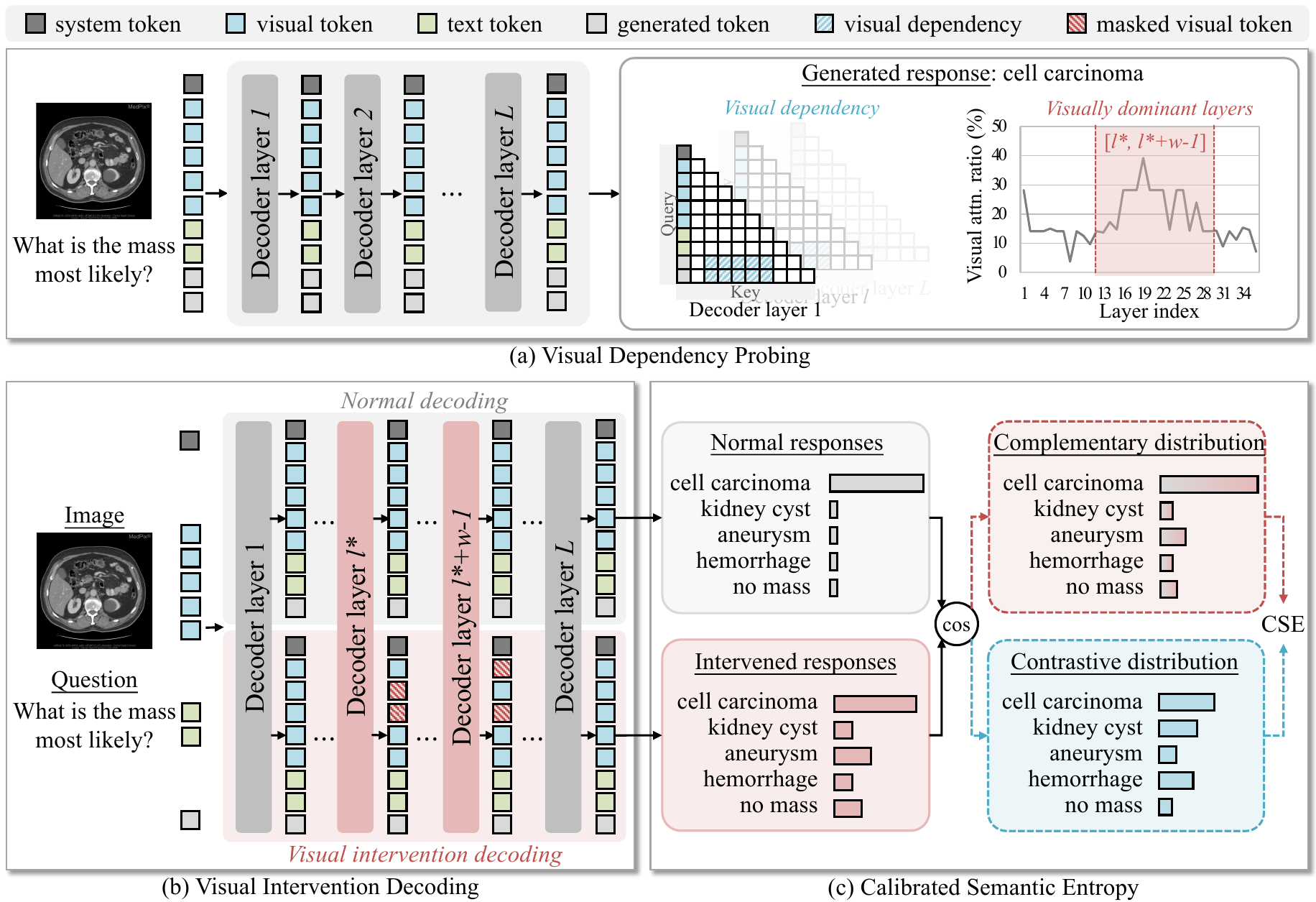}
    \caption{Framework of VIHD. VIHD comprises (a) visual dependency probing, (b) visual intervention decoding, and (c) calibrated semantic entropy.}
    \label{fig:framework}
\end{figure}

As illustrated in Fig.~\ref{fig:framework}, VIHD detects hallucinations via fine-grained and targeted visual interventions. It comprises three components: (1) \textit{Visual Dependency Probing}, which pinpoints visually dominant decoder layers for targeted intervention; (2) \textit{Visual Intervention Decoding}, which performs attention-guided token masking to produce intervened responses; (3) \textit{Calibrated Semantic Entropy}, which calibrates semantic distribution based on normal and intervened responses and quantifies semantic entropy as a hallucination signal.

\subsection{Visual Dependency Probing (VDP)}
Given an input image $x_v$ and a textual query $x_q$, a medical MLLM parametrized with $\theta$ generates a response $r = (r_1, \dots, r_T)$ autoregressively, where $T$ is the sequence length. The probability of generating response $r$~\cite{leng2024mitigating} is factorized as:
\begin{equation}
    p_\theta(r|x_v,x_q) = \prod_{t=1}^T p_{\theta}(r_t \mid x_v, x_q, r_{<t}).
\end{equation}
where $r_t$ is the $t$-th generated token and $r_{<t}$ is the preceding sequence.

Hallucinations often arise when tokens are generated primarily from linguistic priors without sufficient visual evidence. To quantify visual dependency during decoding, we first analyze the attention interactions between generated tokens and the input sequence. Let $\mathbf{Q}_r^{(l)}\in\mathbb{R}^{T\times d}$ denote the query matrix of generated tokens and $\mathbf{K}^{(l)}\in\mathbb{R}^{N\times d}$ represent the key matrix of the input sequence. The attention matrix $\mathbf{A}^{(l,h)}\in \mathbb{R}^{T\times N}$ for the $h$-th head is computed as:
\begin{equation}
    \mathbf{A}^{(l,h)} = \text{softmax}\left(\frac{\mathbf{Q}_{r}^{(l,h)} {\mathbf{K}^{(l,h)}}^\top}{\sqrt{d_h}}\right), \quad h \in \{1, \dots, H\},
\end{equation}
where $d_h=\frac{d}{H}$ denotes the head dimension. We then quantify the layer-wise visual dependency $s^{(l)}$ by aggregating the attention weights assigned to visual tokens (indexed by $\mathcal{V}$) across all heads and generation tokens:
\begin{equation}
s^{(l)} = \frac{1}{H \cdot T} \sum_{h=1}^H \sum_{t=1}^T \sum_{v \in \mathcal{V}} A^{(l,h)}_{t,v}.
\end{equation}
Finally, we localize a window of $w$ consecutive layers with the strongest visual dependency $[l^*,\ {l^*{+}w{-}1}]$. The optimal starting layer $l^*$ is defined as:
\begin{equation}
l^* = \operatorname*{argmax}_l \left( \frac{1}{w}\sum_{i=0}^{w-1} s^{(l+i)} \right).
\end{equation}

\subsection{Visual Intervention Decoding (VID)}
After identifing visually dominant layers via VDP, we propose the visual intervention decoding algorithm to generate the intervened response $r'$. Given a visual input $x_v$ and a textual query $x_q$, VID dynamically intervens the decoding process by applying fine-grained and targeted visual token masking. For each visually dominant layer $l$, we calculate the attention weights between the generated token and visual tokens at each generation step $t$.
\begin{equation}
    \mathbf{A}_{t,v}^{(l)}=\frac{1}{H}\sum_{h=1}^H \mathbf{A}_{t,v}^{(l,h)},\quad(l\in[l^*,l^*{+}w{-}1]).
\end{equation}
We then mask the top-$\rho$ fraction of visual tokens with the highest attention weight $\mathbf{A}_{t,v}^{(l)}$ to suppress their contribution. The intervened response $r'$ is then generated conditioned on the perturbed visual representation. The probability of generating $r'$ is formulated as:
\begin{equation}
    p_{\theta}(r')=\prod_{t=1}^T p_{\theta}(r_t'|x_v,x_q, r_{<t}').
\end{equation}

\subsection{Calibrated Semantic Entropy (CSE)}
To enhance hallucination detection, we compute semantic entropy~\cite{farquhar2024detecting} over a calibrated semantic distribution. Specifically, we perform $M$ independent sampling runs to obtain two sets of responses: $M$ normal responses $\{r^{(i)}\}_{i=1}^M$ and $M$ visual-intervened responses $\{r'^{(i)}\}_{i=1}^M$. Then, we utilize DeBERTa-Large-MNLI~\cite{hedeberta} to cluster them into $N$ semantic-equivalence classes~\cite{kuhnsemantic}, yielding a normal semantic distribution $P(s|x_v,x_q)$ and a visual-intervened one $P(s'|x_v,x_q)$. 

Crucially, visual intervention impacts vary by query granularity. General queries (\textit{e.g.}, imaging modalities) remain robust to token masking, while detailed ones are relatively susceptible to semantic shifts.
Accordingly, we propose an adaptive fusion strategy based on distribution alignment. When the intervened distribution aligns closely with the original, we treat token masking as a weak perturbation and apply \textit{complementary fusion} to reinforce shared semantics. Conversely, we adopt \textit{contrastive fusion} to suppress masking-induced hallucinations. Therefore, the calibrated semantic distribution is formulated as: 
\begin{equation}
\begin{aligned}
P_c(s,s')=
\begin{cases}
\sigma \left(
\frac{1}{1+\alpha} P(s|x_v,x_q) +\frac{\alpha}{1+\alpha}P(s'|x_v,x_q) \right),&\text{if } \cos(P(s),P(s')){\ge}\tau, \\
\sigma \left((1+\alpha) P(s|x_v,x_q) - \alpha P(s'|x_v,x_q)\right), &\text{otherwise}.
\end{cases}
\end{aligned}
\end{equation}
where $\alpha$ denotes the fusion factor, $\sigma(\cdot)$ is the softmax function, and $\tau$ is the similarity threshold. Finally, we calculate the calibrated semantic entropy to quantify hallucination severity. 
\begin{equation}
    CSE = -\sum P_c(s,s')\cdot\log P_c(s,s')
\end{equation}
A higher calibrated semantic entropy indicates significant semantic dispersion, serving as a robust indicator of hallucination.

\section{Experiments}
\subsection{Experimental Settings}
\textbf{Datasets. }
We conduct hallucination detection on the test split of three medical VQA datasets covering multiple imaging modalities (CT, MRI, and X-ray): VQA-RAD~\cite{lau2018dataset}, SLAKE~\cite{liu2021slake}, and VQA-Med-2019~\cite{ImageCLEFVQA-Med2019}. \textbf{VQA-RAD} contains 451 test samples with 200 open-ended and 251 close-ended questions. \textbf{SLAKE} contains 1061 test samples in English with 645 open-ended and 416 close-ended questions. \textbf{VQA-Med-2019} comprises 500 test samples with 436 open-ended and 64 close-ended questions. 

\noindent\textbf{Backbones. }
We perform hallucination detection on two advanced medical MLLMs, Hulu-4B~\cite{jiang2025hulu} and Lingshu-7B~\cite{xu2025lingshu}, for a robust assessment across diverse architectures.

\noindent\textbf{Evaluation Metrics. }
Following~\cite{liao2025vision}, we employ the GREEN model~\cite{ostmeier2024green} to generate hallucination labels. The GREEN model identifies matched findings and clinical errors in generated responses and calculates the GREEN score as $\frac{\#matched}{\#matched+\#errors}$. Samples with $\text{GREEN}<1.0$ are labeled as hallucinated, while others ($\text{GREEN}=1.0$) are non-hallucinated. Performance is quantified using two metrics: (1) \textit{Area Under the ROC Curve (AUC)}, which measures the ranking capability to differentiate hallucinated from non-hallucinated responses; and (2) \textit{Area under the GREEN Curve (AUG)}~\cite{liao2025vision}, which evaluates the alignment between estimated hallucination scores and factual correctness.

\noindent\textbf{Implementation Details. }
During hallucination label construction, we set the LLM sampling temperature to $0.1$ to produce stable primary responses. During hallucination detection, we increase the temperature to $1.0$ and perform $M=10$ independent sampling runs following~\cite{liao2025vision}. We adopt Nucleus sampling~\cite{holtzmancurious}, employ a sliding window of width $w=\frac{L}{2}$ for visual dependency probing, and mask $\rho=10\%$ of high-attention visual tokens in visual intervention decoding. For semantic distribution calibration, we set the cosine similarity threshold to $\tau=0.95$ and the calibration coefficient to $\alpha =1.0$. All experiments are conducted on a single NVIDIA A100 GPU with 80 GB of memory.

\noindent\textbf{Comparison Methods. }
We compared VIHD with seven state-of-the-art introspective methods: (1) four token uncertainty-based methods: AvgProb, MaxProb, AvgEnt, and MaxEnt~\cite{li2024reference}, (2) one consistency-based method: RadFlag~\cite{zhang2024radflag}, and (3) two semantic uncertainty-based methods: SE~\cite{farquhar2024detecting} and VASE~\cite{liao2025vision}.

\begin{table}[t]
    \centering
    \caption{Performance comparison of VIHD and seven competing methods.}
    \setlength{\tabcolsep}{0.3mm}
    \resizebox{\linewidth}{!}{
    \begin{tabular}{l|cc|cc|cc|cc|cc|cc}
        \bottomrule
        \multicolumn{1}{c|}{\multirow{3}{*}{Method}} & \multicolumn{4}{c|}{VQA-RAD} & \multicolumn{4}{c|}{SLAKE} & \multicolumn{4}{c}{VQA-Med-2019}\\
        \cline{2-13}
        & \multicolumn{2}{c|}{Open-Ended} & \multicolumn{2}{c|}{All} & \multicolumn{2}{c|}{Open-Ended} & \multicolumn{2}{c|}{All} & \multicolumn{2}{c|}{Open-Ended} & \multicolumn{2}{c}{All}\\
        & AUC & AUG & AUC & AUG & AUC & AUG & AUC & AUG & AUC & AUG & AUC & AUG\\
        \hline
        \rowcolor{gray!20}\multicolumn{13}{c}{Hulu-4B}\\
        \hline
        AvgProb & 62.01 & 53.68 & 60.13 & 67.02 & 60.30 & 68.24 & 59.13 & 72.59 & 70.48 & 51.56 & 67.90 & 55.00 \\
        MaxProb & 62.52 & 53.54 & 61.17 & 66.98 & 60.37 & 68.21 & 59.26 & 72.62 & 71.45 & 51.86 & 69.45 & 55.39 \\
        AvgEnt  & 61.15 & 53.56 & 59.32 & 66.88 & 60.39 & 68.26 & 59.03 & 72.56 & 70.43 & 51.68 & 67.72 & 55.00 \\
        MaxEnt  & 62.49 & 53.71 & 61.12 & 66.98 & 60.45 & 68.22 & 59.24 & 72.61 & 71.67 & 52.74 & 69.64 & 55.45 \\
        RadFlag & 50.00 & 49.30 & 50.00 & 61.26 & 50.00 & 59.02 & 50.00 & 68.12 & 50.00 & 21.74 & 50.00 & 22.84 \\
        SE      & 50.89 & 51.77 & 49.48 & 62.24 & 49.29 & 59.02 & 49.30 & 66.89 & 47.53 & 19.58 & 47.42 & 21.43 \\
        VASE    & 76.98 & 69.09 & 66.23 & 71.51 & 74.41 & 79.12 & 70.31 & 81.27 & 75.09 & 54.30 & 72.99 & 57.41 \\
        \hline
        \rowcolor{blue!10}VIHD & $\bm{83.13}$ & $\bm{72.01}$ & $\bm{78.23}$ & $\bm{80.82}$ & $\bm{79.94}$ & $\bm{81.11}$ & $\bm{78.91}$ & $\bm{84.96}$ & $\bm{78.65}$ & $\bm{59.64}$ & $\bm{78.10}$ & $\bm{60.39}$\\
        \blue{$\mathrm{\Delta}$} & \blue{+6.15} & \blue{+2.92} & \blue{+12.00} & \blue{+9.31} & \blue{+5.53} & \blue{+1.99} & \blue{+8.60} & \blue{+3.69} & \blue{+3.56} & \blue{+5.34} & \blue{+5.11} & \blue{+2.98} \\
        \hline
        \rowcolor{gray!20}\multicolumn{13}{c}{Lingshu-7B}\\
        \hline
        AvgProb & 56.23 & 39.63 & 54.72 & 59.29 & 53.94 & 75.76 & 53.30 & 77.39 & 55.61 & 35.54 & 55.11 & 36.81 \\
        MaxProb & 56.21 & 39.67 & 54.71 & 59.30 & 53.93 & 75.75 & 53.30 & 77.38 & 55.59 & 35.54 & 55.12 & 36.78 \\
        AvgEnt  & 56.28 & 39.73 & 54.80 & 59.34 & 53.97 & 75.82 & 53.34 & 77.42 & 55.68 & 35.63 & 55.11 & 36.90\\
        MaxEnt  & 56.27 & 39.74 & 54.74 & 59.33 & 53.99 & 75.82 & 53.37 & 77.44 & 55.64 & 35.59 & 55.15 & 36.82\\
        RadFlag & 53.27 & 36.33 & 52.94 & 58.94 & 57.07 & 77.14 & 55.46 & 78.25 & 57.30 & 35.91 & 56.04 & 37.85 \\
        SE      & 50.37 & 35.93 & 50.92 & 55.27 & 52.72 & 74.13 & 52.45 & 76.44 & 54.27 & 34.69 & 53.96 & 35.96 \\
        VASE    & 71.45 & 48.02 & 69.62 & 69.35 & 72.37 & 85.06 & 70.56 & 84.52 & 69.57 & 56.77 & 68.85 & 60.80 \\
        \hline
        \rowcolor{blue!10} VIHD    & $\bm{73.13}$ & $\bm{55.84}$ & $\bm{71.58}$ & $\bm{70.86}$ & $\bm{75.36}$ & $\bm{88.61}$ & $\bm{72.95}$ & $\bm{86.49}$ & $\bm{74.71}$ & $\bm{64.11}$ & $\bm{74.18}$ & $\bm{64.54}$\\
        \blue{$\mathrm{\Delta}$} & \blue{+1.68} & \blue{+7.82} & \blue{+1.96} & \blue{+1.51} & \blue{+2.99} & \blue{+3.55} & \blue{+2.39} & \blue{+1.97} & \blue{+5.14} & \blue{+7.34} & \blue{+5.33} & \blue{+3.74} \\

        \toprule
    \end{tabular}
    }
    \label{tab:main}
\end{table}

\subsection{Results}
Table~\ref{tab:main} presents the hallucination detection performance of VIHD against seven competing methods across three medical VQA benchmarks. 
The best performances are highlighted in \textbf{bold} and improvements over the second-best one are shown in \blue{blue}. As reported in Table~\ref{tab:main}, VIHD consistently outperforms competing methods by a notable margin across all VQA datasets and medical MLLMs. On \textbf{Hulu-4B}, it improves the average AUC by 8.57\% for all questions and 5.08\% for open-ended questions. Notably, the most substantial gain is observed on VQA-RAD, where VIHD achieves a 12\% AUC increase across all questions.
On the more capable \textbf{Lingshu-7B} backbone, VIHD maintains its advantage with average AUC improvements of 3.22\% overall and 3.27\% on open-ended questions, with a substantial 5.33\% gain on VQA-Med-2019. These results validate VIHD's effectiveness in detecting hallucinations across diverse medical imaging modalities and question types. The consistent improvements over VASE across both model capacities and dataset characteristics underscore the efficacy of internal fine-grained visual intervention in hallucination detection. Furthermore, VIHD requires 11.59s per sample at inference, comparable to VASE (11.55s/sample), demonstrating that performance improvements are achieved without additional computational cost.

\subsection{Ablation Study and Further Analysis}
\noindent\textbf{Ablation Study. } Table~\ref{tab:ablation} presents the component-wise ablation study on VQA-RAD and VQA-Med-2019 using Hulu-4B. Without any component, the baseline, Semantic Entropy~\cite{farquhar2024detecting}, yields overall AUC scores of 49.48\% and 47.42\% on VQA-RAD and VQA-Med-2019, respectively. The introduction of visual intervention decoding (VID) substantially improves AUC by 18.37\% and 27.35\%, demonstrating the effectiveness of targeted visual intervention. Augmenting VID with calibrated semantic entropy (CSE) further boosts performance by 8.45\% and 1.79\%. Finally, incorporating visual dependency probing (VDP) to focus interventions at visually dominant layers leads to the best overall performance. These results validate the effectiveness of each proposed component.
\begin{table}[tbp]
    \centering
    \setlength{\tabcolsep}{1.5mm}
    \caption{Ablation study on VQA-RAD and VQA-Med-2019 using Hulu-4B. }
    \begin{tabular}{ccc|cc|cc|cc|cc}
        \bottomrule
        \multicolumn{3}{c|}{Module} & \multicolumn{4}{c|}{VQA-RAD} & \multicolumn{4}{c}{VQA-Med-2019}\\
        \hline
        \multirow{2}{*}{VDP} & \multirow{2}{*}{VID} & \multirow{2}{*}{CSE} & \multicolumn{2}{c|}{Open-Ended} & \multicolumn{2}{c|}{All} & \multicolumn{2}{c|}{Open-Ended} & \multicolumn{2}{c}{All}\\
         & & &  AUC & AUG & AUC & AUG & AUC & AUG & AUC & AUG\\
        \hline
        \ding{55} & \ding{55} & \ding{55} & 50.89 & 51.77 & 49.48 & 62.24 & 47.53 & 19.58 & 47.42 & 21.43\\
        \ding{55} & \ding{51} & \ding{55} & 79.66 & 71.79 & 67.85 & 73.96 & 76.23 & 56.54 & 74.77 & 59.23\\
        \ding{55} & \ding{51} & \ding{51} & 79.97 & 71.65 & 76.30 & 80.35 & 77.64 & 58.80 & 76.56 & 59.73\\
        \ding{51} & \ding{51} & \ding{51} & $\bm{83.13}$ & $\bm{72.01}$ & $\bm{78.23}$ & $\bm{80.82}$ & $\bm{78.65}$ & $\bm{59.64}$ & $\bm{78.10}$ & $\bm{60.39}$\\ 
        \toprule
    \end{tabular}
    \label{tab:ablation}
\end{table}
\begin{figure}[t]
    \begin{minipage}[t]{0.49\textwidth}
        \centering
        \includegraphics[width=\textwidth]{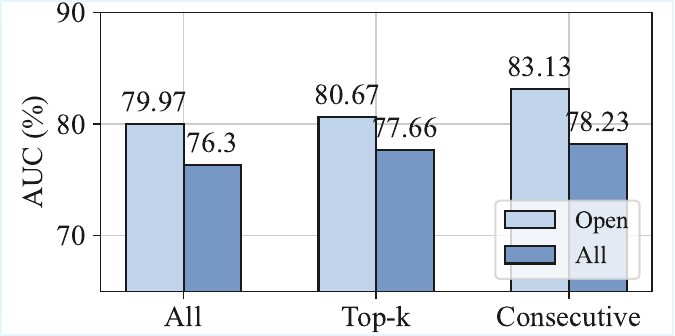}
        \caption{Variants of VDP.} 
        \label{fig:VDP}
    \end{minipage}
    \hfill 
    \begin{minipage}[t]{0.49\textwidth}
        \centering
        \includegraphics[width=\textwidth]{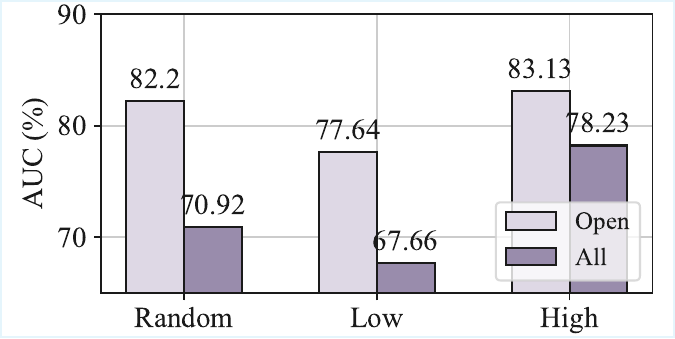}
        \caption{Variants of visual token masking.} 
        \label{fig:VTM}
    \end{minipage}
\end{figure}

\noindent\textbf{Variants of Visual Dependency Probing (VDP). }
\begin{figure}[t]
    \centering
    \begin{minipage}[t]{0.49\textwidth}
        \strut\\[-\baselineskip]
        \centering
        \includegraphics[width=\linewidth]{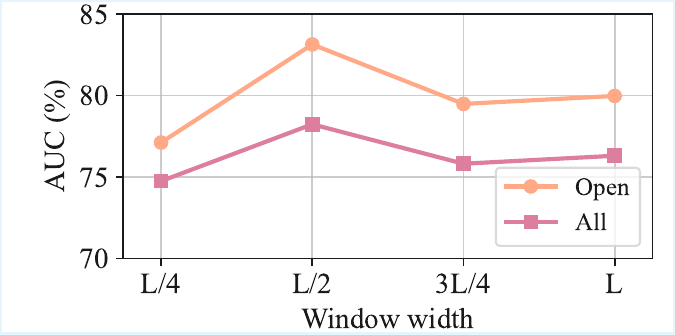}
        \captionof{figure}{Effect of sliding window width.}
        \label{fig:win_width}
        
        \includegraphics[width=\linewidth]{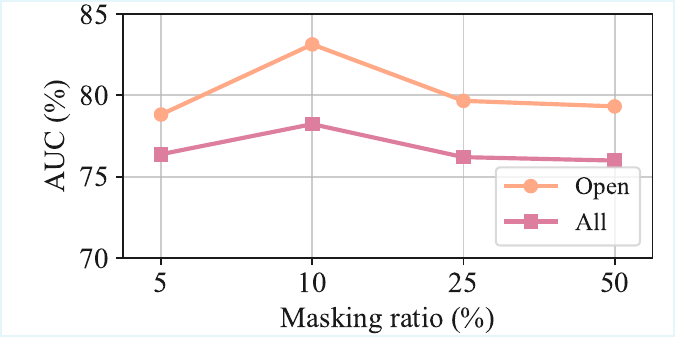}
        \captionof{figure}{Effect of masking ratio.}
        \label{fig:mask_ratio}
    \end{minipage}
    \hfill
    \begin{minipage}[t]{0.49\textwidth}
        \centering
        \captionof{table}{Performance comparison on VQA-RAD using Hulu-4B with Top-K sampling.}
        \label{tab:topk}
        \begin{tabular}{l|cc|cc}
        \bottomrule
        \multicolumn{1}{c|}{\multirow{3}{*}{Method}} & \multicolumn{4}{c}{VQA-RAD}\\
        \cline{2-5}
        & \multicolumn{2}{c|}{Open-Ended} & \multicolumn{2}{c}{All}\\
        & AUC & AUG & AUC & AUG\\
        \hline
        AvgProb & 65.06 & 53.51 & 60.68 & 69.58 \\
        MaxProb & 65.89 & 53.24 & 61.73 & 69.79 \\
        AvgEnt  & 63.61 & 51.83 & 59.88 & 69.30 \\
        MaxEnt  & 65.06 & 53.41 & 61.73 & 69.85 \\
        RadFlag & 50.00 & 44.77 & 50.00 & 62.54 \\
        SE      & 48.21 & 42.30 & 48.56 & 58.64 \\
        VASE    & 76.84 & 67.50 & 70.40 & 75.48 \\
        \hline
        \rowcolor{blue!10} VIHD (Ours) & 80.56 & 71.85 & 77.07 & 80.25\\
        \blue{$\mathrm{\Delta}$} & \blue{+3.72} & \blue{+4.35} & \blue{+6.67} & \blue{+4.77}\\
        \toprule
        \end{tabular}
    \end{minipage}
\end{figure}
As shown in Fig.~\ref{fig:VDP}, we evaluate three layer selection strategies for visual dependency probing: (1) \textit{all-layer selection}, which applies token masking across all decoder layers; (2) \textit{attention-based selection}, which identifies isolated layers with peak visual dependency; and (3) \textit{consecutive attention-based selection}, which selects a sequential span of high cumulative dependency to enforce continuity. Compared to the all-layer selection, the attention-based strategy yields a 1.36\% AUC gain on VQA-RAD, highlighting the necessity of targeting visually dominant layers. Enforcing layer continuity further improves AUC by 0.57\%. Therefore, we adopt the consecutive attention-based strategy as our default.

\noindent\textbf{Variants of Visual Token Masking. }
We evaluate three token masking strategies within visual intervention decoding: (1) \textit{random masking}, (2) \textit{low-attention masking}, and (3) \textit{high-attention masking}. As shown in Fig.~\ref{fig:VTM}, high-attention masking outperforms random masking by 7.41\% AUC on VQA-RAD, which demonstrates that targeted perturbations of salient visual tokens facilitate effective semantic calibration. Conversely, low-attention masking acts merely as a subtle augmentation with negligible impact, thus leading to limited detection. Therefore, we adopt high-attention masking in visual intervention decoding.

\noindent\textbf{Effect of Sliding Window Width. }
We also analyze the impact of sliding window width $w=\{\frac{L}{4}, \frac{L}{2},\frac{3L}{4}, L\}$ on intervention efficacy. As illustrated in Fig.~\ref{fig:win_width}, performance peaks at $w=\frac{L}{2}$. Narrower windows $\frac{L}{4}$ restrict contextual propagation of perturbations, limiting their discriminative power, while wider windows $\frac{3L}{4}$ introduce excessive noise by intervening semantically heterogeneous layers. The $\frac{L}{2}$ configuration strikes an optimal balance, enabling localized yet contextually coherent perturbation propagation. We adopt $w=\frac{L}{2}$ for all experiments.

\noindent\textbf{Effect of Masking Ratio. }
We investigate the impact of masking ratio $\rho\in\{5\%, 10\%, 25\%, 50\%\}$ on VQA-RAD. As shown in Fig.~\ref{fig:mask_ratio}, insufficient masking $(\rho=5\%)$ yields weak perturbations that fail to calibrate semantic distribution, while aggressive masking $(\rho>25\%)$ corrupts visual semantics and degrades calibration fidelity. In contrast, the ratio of $10\%$ achieves the optimal detection performance by introducing discriminative yet non-destructive interventions. Therefore, we fix $\rho=10\%$ throughout experiments.

\noindent\textbf{Effect of Sampling Strategy. }
To assess robustness under diverse decoding strategies, we evaluate all methods using Top-K sampling (K=50)~\cite{fan2018hierarchical} on VQA-RAD. As shown in Table~\ref{tab:topk}, VIHD maintains superior performance across both AUC and AUG metrics, surpassing the strongest baseline (VASE) by 6.67\% in AUC and 4.77\% in AUG overall. These results validate the effectiveness of visual intervention for hallucination detection across multiple sampling paradigms.

\section{Conclusion}
We have proposed \textbf{V}isual \textbf{I}ntervention-based \textbf{H}allucination \textbf{D}etection (VIHD), a training-free method that exploits fine-grained visual intervention to detect hallucinations. VIHD identifies visually dominant decoder layers, performs visual intervention decoding to derive calibrated semantic entropy as a hallucination score. Extensive experiments on three medical VQA datasets and two MLLMs have demonstrated the superiority of VIHD in hallucination detection.

\begin{credits}
\subsubsection{\ackname} 
This work was supported by the Medical Research Future Fund (MRFF) under Grant NCRI000074.
\end{credits}

\bibliographystyle{splncs04}
\bibliography{Paper-4543}

\end{document}